# Deep Morphological Simplification Network (MS-Net) for Guided Registration of Brain Magnetic Resonance Images

Dongming Wei, Zhengwang Wu, Gang Li, Xiaohuan Cao, Dinggang Shen*, Fellow, IEEE and Qian Wang*, Member, IEEE

*Abstract*—Objective: Deformable brain MR image registration is challenging due to large inter-subject anatomical variation. For example, the highly complex cortical folding pattern makes it hard to accurately align corresponding cortical structures of individual images. In this paper, we propose a novel deep learning way to simplify the difficult registration problem of brain MR images. Methods: We train a morphological simplification network (MS-Net), which can generate a "simple" image with less anatomical details based on the "complex" input. With MS-Net, the complexity of the fixed image or the moving image under registration can be reduced gradually, thus building an individual (simplification) trajectory represented by MS-Net outputs. Since the generated images at the ends of the two trajectories (of the fixed and moving images) are so simple and very similar in appearance, they are easy to register. Thus, the two trajectories can act as a bridge to link the fixed and the moving images, and guide their registration. Results: Our experiments show that the proposed method can achieve highly accurate registration performance on different datasets (i.e., NIREP, LPBA, IBSR, CUMC, and MGH). Moreover, the method can be also easily transferred across diverse image datasets and obtain superior accuracy on surface alignment. Conclusion and Significance: We propose MS-Net as a powerful and flexible tool to simplify brain MR images and their registration. To our knowledge, this is the first work to simplify brain MR image registration by deep learning, instead of estimating deformation field directly.

## I. Introduction

DEFORMABLE image registration [1], [2] aims to estimate the deformation field, following which the moving image can be warped to the space of the fixed image. This technique plays an important role in medical image analysis, as it can help build anatomical correspondences across images and facilitate the subsequent analysis. Whereas registration is often perceived as an optimization problem, the deformation field needs to be optimized iteratively, with certain smoothness regularization, to maximize the similarity between the fixed and the moving images. Commonly used methods for brain magnetic resonance (MR) image registration include AIR [3], ART [4], SyN [5], HAMMER [6], Demons [7]–[9], SPM [10], DRAMMS [11], DROP [12], CC/MI/SSD-FFD [13], FNIRT [14], LDDMM [15], etc. Although comprehensive comparisons of these methods are reported in [16], [17], it is still difficult to assert the best algorithm for a certain application especially when dealing with diverse datasets.

The large anatomical variation across different images is a great challenge to image registration. In brain MR images, the cortical folding patterns are known to be complex with high inter-subject variation. Whereas imaging-based studies require highly accurate alignment of the corresponding neuroanatomies across different subjects, most existing methods struggle in estimating deformation fields to register tiny structures (e.g., cortical areas) precisely. For example, one may evaluate the overlap ratio between the same anatomical structures of the fixed image and the warped moving image as a metric of the registration quality [18]. Although a reduced smoothness constraint of the deformation field may increase the overlap metric, the topology-preserving property of the deformation fields would then be at high risk to be destroyed, leading to a possible failure to the entire registration task. Thus, a high-performance registration method, which could consistently work well for different datasets and tasks with minimal parameter tuning, is of great interest to the community.

To address the concern of large anatomical variation, several works have introduced intermediate images into the deformation pathway between the fixed and the moving images [19], [20]. A manifold is often instantiated to account for the distribution of the imaging data. Then, the very long pathway connecting two images that are far away on the manifold is divided into several short segments by the intermediate images, each of which corresponds to an easier-to-estimate deformation field. However, it is non-trivial to create the manifold. The imaging data are high-dimensional, implying that a sufficient number of (intermediate) images is necessary to model the complex distribution of the image population. Meanwhile, a

Dongming Wei and Qian Wang are in the Institute for Medical Imaging Technology, School of Biomedical Engineering, Shanghai Jiao Tong University, 200030, Shanghai, China (email: wang.qian@sjtu.edu.cn).

Zhengwang Wu, Gang Li and Dinggang Shen are with the Department of Radiology and BRIC, University of North Carolina at Chapel Hill, Chapel Hill, North Carolina 27599, USA. Dinggang Shen is also with Department of Brain and Cognitive Engineering, Korea University, Seoul 02841, Republic of Korea (email: dgshen@med.unc.edu).

Xiaohuan Cao is with Shanghai United Imaging Intelligence Co., Ltd., Shanghai, China.

global image similarity metric is needed by the manifold, yet the metric is challenging to design and often fails to describe local anatomical variation effectively [21].

In last decades, machine learning has become a frequently used tool to image registration. In particular, convolution neural network (CNN) is recently employed to directly predict the deformation field from a pair of fixed/moving images in [22], [23], where the ground truth for training is acquired by SyN and Demons. The initial momenta for LDDMM can also be predicted by CNN as in [24]. Moreover, it is shown that the deformation field can be predicted from the input images through a deep network trained without supervision [25]–[28]. That is, the image similarity metric and the smoothness regularization can jointly guide the training of the network in back-propagation. The adversarial strategy is also used to impose the regularization such that the local minima in registration can be better avoided [29]. Although registration can be solved in black-box by the powerful computation capability of deep learning, most previous works fail to consider the high complexity of the image manifold. The issue of large anatomical variation is still challenging to many brain MR image registration applications.

In this paper, we propose a novel deep learning way to simplify brain MR image registration. Specifically, we train CNNs to reduce the anatomical complexity of the fixed and the moving images (i.e., cortical folding pattern), which is a bottleneck to the registration task. We derive a trajectory from fixed/moving image, which consists of a series of images with gradually-reduced anatomical complexity. As the two trajectories approach to the ends, the anatomical complexity of the simplified fixed/moving images becomes low, implying that the two images become similar with each other in the simplified morphological space. The two simplified images are then easier to be registered compared to the case of directly registering the original fixed and moving images. In this way, the two trajectories generated by deep learning can act as a bridge to link the fixed/moving images. By composing multiple deformable registration tasks along the two trajectories, the moving image can finally be registered with the fixed image accurately and reliably.

Our method is unique as it breaks the barrier of the complex image manifold in deformable registration of brain MR images, and provides intermediate guidance through deep learning for the first time. To effectively simplify the complexity of brain MR images, we train the morphological simplification networks (MS-Nets) particularly. MS-Nets are able to generate a set of T1 images given an input image. The generated images are gradually simplified in terms of the cortical folding patterns, while no segmentation, parcellation, or cortical surface reconstruction is needed for a test image. To our knowledge, our method is the first to simplify brain MR image registration via deep learning, instead of estimating the deformation field directly as a black-box. Moreover, our experiments show that the MS-Nets trained with a certain dataset are robust to transfer to other new datasets, making our method highly adaptable to many clinical applications.

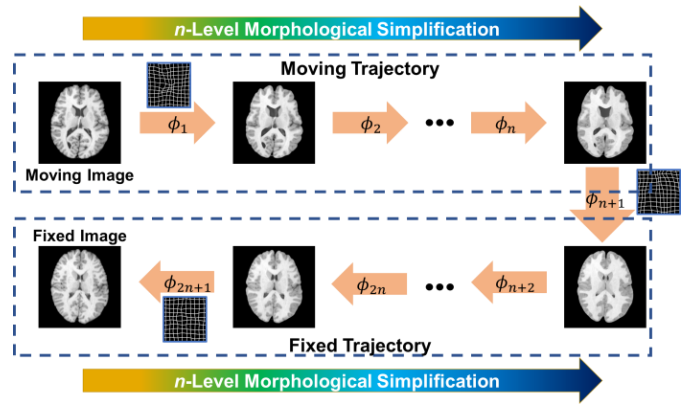

Fig. 1. Illustration of the pipeline of the proposed method. The fixed and the moving images reduce their anatomical complexity gradually through n-level morphological simplification process by deep learning, which results in the fixed/moving (image) trajectories, respectively. At the ends of the two trajectories, the fixed and the moving images become similar in appearance, such that their registration can be easily completed. Finally, the registration between the moving and the fixed images can be attained by concatenating multiple deformation fields (i.e., $\{\phi_i | i = 1, \cdots, 2n + 1\}$), each of which is denoted by an arrow.

## II. METHOD

We propose to simplify the registration of brain MR images by deep learning. The pipeline of our method is shown in Fig. 1. In particular, we train the MS-Nets to reduce the anatomical complexity, and generate the trajectories for the fixed/moving images. The anatomical complexity is gradually reduced along each trajectory, while the images at the ends of the fixed/moving trajectories become simple and similar, implying that they are easy to be registered in the simplified morphological space. In this way, we can follow the fixed/moving (image) trajectories and decompose the original complex registration problem into several easy ones.

### A. Morphological Simplification Network (MS-Net)

The key point of our method relies on the gradual reduction of the anatomical complexity in brain MR images. It is known that the cortex is highly folded in human brain. To acquire more accurate alignment of the anatomical structures, high-order features and sophisticated constraints derived from brain tissue segmentation are shown to be effective [6], [30]. Recently, Zhang et al. proposed to use the smoothed cortical surface with reduced complexity to guide the registration of the 3D brain volumes [31]. However, it is non-trivial to get high-quality tissue segmentation especially when multi-center data is considered – sometimes even expert editing of tissue segmentation is necessary. To this end, we propose to complete morphological simplification of brain MR images in the intensity space by deep learning, without need of any segmentation or surface reconstruction to the test image.

*Training data preparation.* We aim to train a morphological simplification network (MS-Net) to reduce the complexity in brain MR images. Specifically, the input to MS-Net is a complex image and the corresponding output is a simple one. To prepare the "ground-truth" data to supervise the training of MS-Net, we first segment the tissues of grey matter (GM) and white matter (WM) of the "complex" image, and reconstruct the inner/outer cortical surfaces. Then, we apply Laplacian smoothing upon the cortical surface meshes. With $x_i$ representing the location of the i-th vertex of the mesh of the inner/outer cortical surface, its new coordinate after smoothing is

$$x_i := \sum_{\substack{j \in \mathcal{N}_i \\ j \neq i}} w_{ij}(x_j - x_i), \quad (1)$$

where $x_j$ is the neighbor of the vertex $x_i$, $w_{ij} = 1/m$ indicates the weight, and $m$ is the size of the neighborhood. Next, we convert the smoothed surfaces back to the binary volumes of GM/WM tissue labels. To avoid volume shrinkage, we follow the strategy in [32] to assure that the tissue label volumes are not changed before/after smoothing. Next, for each image, we register the binary tissue label volumes before/after smoothing and generate the deformation field [8]. Finally, we apply the generated deformation field to the "complex" image and get the "simple" image.

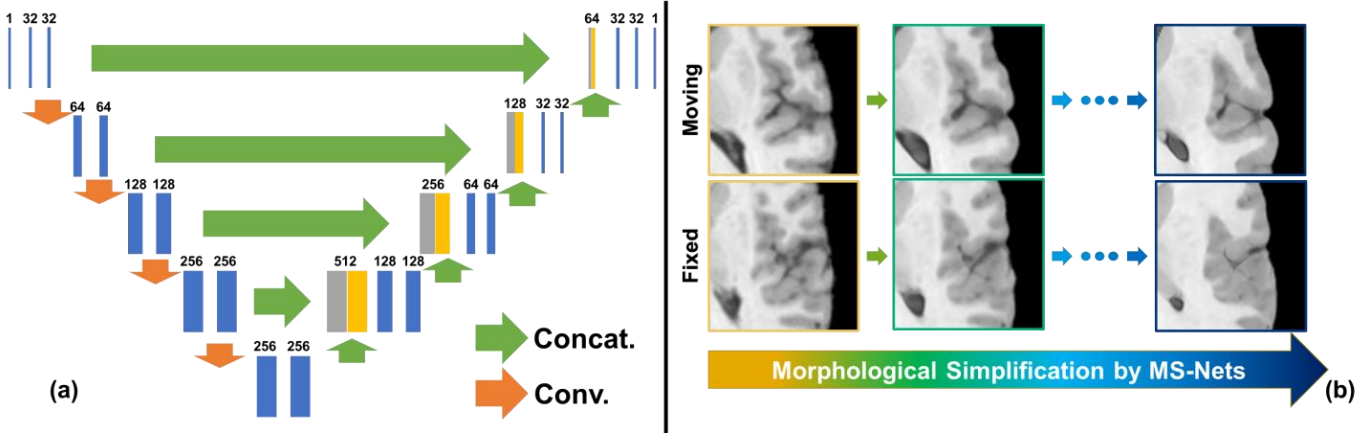

Fig. 2. (a) The architecture of MS-Net. Each blue box indicates a multi-channel feature map. The number of the channels is denoted with the box. The kernel size is 3×3×3 for all layers. Grey and yellow boxes represent the copied feature maps. (b) The two example images and their simplified outputs by MS-Nets: From left to right, the complexity of the original fixed/moving image is reduced gradually.

*Network configuration and training.* The architecture of MS-Net and the detailed configurations can be found in Fig. 2(a). For easy illustration, we show the 3D layers inside the network by 2D boxes in the figure, while the number above each box indicates the channel number after convolution and concatenation. Each layer in the MS-Net is a 3D layer without pooling. The kernel size is 3×3×3 and the stride is 1. Zero padding is adopted to keep the sizes of the feature maps and also make the output the same as the input through the MS-Net.

We train the MS-Net in a patch-by-patch way. Particularly, we sample 3D cubic patches sized 16×16×16 from the training images. The sampling complies with the probability calculated at the center of each potential patch (denoted by u):

$$p(u) = \frac{|\nabla g_u^x| + |\nabla g_u^y| + |\nabla g_u^z|}{\|\nabla g\|}, \quad (2)$$

where $\nabla g_u^x, \nabla g_u^y$ and $\nabla g_u^z$ are the gradients at $u$ in three directions, and $\|\nabla g\|$ is the gradient norm. In this way, the sampled patches cover the entire brain volumes and pay more attention to the regions of abundant appearance information [23]. To train the MS-Net, we usually extract 20,000 patch samples from each pair of the prepared complex and simple images. The network is trained on an Nvidia Titan X GPU by Keras. The optimizer is Adam with 0.001 as the initial learning rate. For the loss function, we use the sum of the squared differences between the output patch and the ground-truth patch.

*Application of the MS-Net.* In the application stage, the trained network can be directly applied to generate the 3D simple output image from a complex input in the end-to-end way. Concerning the capacity of GPU, we implement to process every 16 axial slices for each test task. The whole test image can then generate its simple version by averaging the results of all test tasks, with two neighboring tasks sharing 8 overlapped slices. In this way, a typical test image sized 256×256×256 is simplified by the MS-Net within ~3.3 seconds. Examples of the fixed/moving images and their simplified outputs are available in Fig. 2(b). Note that several MS-Nets are applied to derive the trajectories for the fixed/moving images in the figure.

### B. Trajectory and its Guidance to Registration

To generate the trajectory where a brain MR image is gradually simplified, we train a sequence of MS-Nets one by one. In particular, there are 7 levels of morphological simplification in our implementation ($n$=7 as in Fig. 1), corresponding to 7 different MS-Nets. Each MS-Net is assigned to generate a corresponding intensity image with adequate smoothing scale, to ensure that the two consecutive images in the trajectory can be similar enough and then easily registered.

After trajectory construction, here we need to calculate the accurate deformation field between the fixed and the moving images. The desired deformation field can be derived by composing multiple deformation fields along the two trajectories, as well as between the ends of them. In particular, we adopt Diffeomorphic Demons [8] to estimate $\phi_i$ ($i = 1, \cdots, 2n + 1$) as in Fig. 1. Each $\phi_i$ is relatively easy to compute, as the decomposed registration always happens between

TABLE I
REGISTRATION ACCURACY EVALUATED ON THE NIREP DATASET.

| Evaluation on GM/WM Tissue Labels | | | |
|---|---|---|---|
| | **Proposed** | **Demons** | **SyN** |
| GM (DSC: %) | **87.06±0.91** | 81.69±1.02 | 81.59±2.26 |
| WM (DSC: %) | **89.60±0.72** | 84.20±0.80 | 83.91±2.10 |
| GM (ASSD: mm) | **0.30±0.05** | 0.43±0.06 | 0.40±0.08 |
| WM (ASSD: mm) | **0.41±0.09** | 0.53±0.09 | 0.53±0.12 |
| **Evaluation on 32 small ROIs (TO: %)** | | | |
| | **Proposed** | **Demons** | **SyN** |
| Overall | **70.78±5.00** | 67.39±6.26 | 66.92±6.96 |

| | **Proposed** | **Demons** | **SyN** | **Proposed** | **Demons** | **SyN** |
|---|---|---|---|---|---|---|
| | Left Hemisphere | | | Right Hemisphere | | |
| Occipital Lobe | 71.81±8.61 | 65.74±8.51 | 66.87±7.83 | 74.23±6.08 | 66.73±7.31 | 69.34±5.88 |
| Cingulate Gyrus | 68.26±9.01# | 68.94±8.60 | 68.00±8.28 | 69.36±7.89# | 69.42±8.00 | 68.67±7.72 |
| Insula Gyrus | 76.74±4.14 | 76.69±4.64 | 75.85±4.68 | 77.66±3.62# | 78.68±4.97 | 78.03±3.39 |
| Temporal Pole | 75.63±11.97# | 76.09±9.90 | 74.88±11.04 | 78.49±8.04 | 77.70±8.02 | 76.59±8.24 |
| Superior Temporal Gyrus | 68.11±7.73 | 66.00±8.12 | 65.69±7.63 | 67.85±8.07 | 64.62±9.03 | 64.50±8.02 |
| Infero Temporal Region | 76.15±5.07 | 72.09±5.04 | 72.36±5.23 | 77.19±5.45 | 71.85±6.46 | 72.65±5.83 |
| Parahippocampal Gyrus | 73.22±5.45 | 72.86±5.74 | 72.01±5.51 | 75.29±5.72# | 75.29±5.70 | 74.46±5.34 |
| Frontal Pole | 74.15±9.55 | 71.49±10.19 | 72.43±9.65 | 72.91±10.56 | 69.84±10.34 | 70.94±9.85 |
| Superior Frontal Gyrus | 71.86±7.18 | 68.20±8.43 | 68.01±7.27 | 72.24±8.48 | 67.20±9.67 | 67.51±8.29 |
| Middle Frontal Gyrus | 70.84±8.95 | 64.49±9.10 | 65.67±8.71 | 67.27±8.64 | 60.17±8.69 | 61.66±8.09 |
| Inferior Gyrus | 66.10±14.28 | 60.44±14.09 | 61.04±14.00 | 67.41±9.98 | 60.03±11.49 | 62.26±9.77 |
| Orbital Frontal Gyrus | 75.97±6.34 | 74.67±6.84 | 73.70±6.29 | 75.16±5.88 | 73.58±6.17 | 72.66±5.57 |
| Precentral Gyrus | 66.31±8.13 | 61.22±8.48 | 62.46±7.28 | 64.11±7.42 | 58.89±8.44 | 59.99±7.33 |
| Superior Parietal Lobule | 67.33±9.30 | 59.07±10.9 | 61.23±9.27 | 66.28±7.85 | 57.94±7.87 | 60.81±7.25 |
| Inferior Parietal Lobule | 69.30±8.16 | 61.73±8.57 | 64.28±7.78 | 68.71±8.95 | 60.18±9.42 | 63.30±8.82 |
| Postcentral Gyrus | 60.70±12.24 | 56.37±12.46 | 55.94±11.79 | 58.17±9.71 | 52.94±9.28 | 52.42±9.34 |

# indicates the ROIs (i.e., Left Cingulate Gyrus, Right Cingulate Gyrus, Right Insula Gyrus, Left Temporal Pole and Right Parahippocampal Gyrus) where our method does not outperform both Demons and SyN significantly at the same time (i.e., $p>0.01$ in paired $t$-tests against either Demons or SyN).

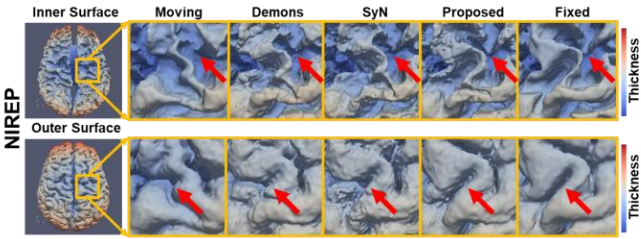

Fig. 3. Visualization of the registration results of the NIREP dataset by Demons, SyN and our proposed method. The images are shown in reconstructed inner (top row) and outer (bottom row) cortical surfaces, and colored in accordance to cortical thickness. Our method shows more accurate surface alignment especially in the regions highlighted by red arrows.

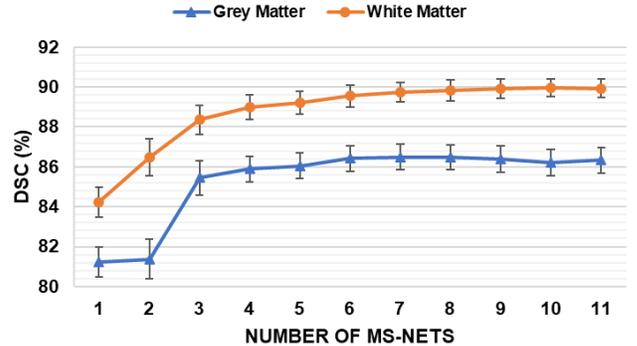

Fig. 4. The mean DSC of GM/WM based on 15 image pairs drawn from the NIREP dataset, with respect to the gradually increasing number of MS-Nets used to derive the fixed/moving sequences.

images of similar appearance. In the end, the deformation field $\phi$ that warps the moving image to the fixed image is obtained by composing all decomposed registration tasks:

$$\phi = \phi_1 \circ \phi_2 \circ \cdots \circ \phi_{2n+1}. \quad (3)$$

## III. EXPERIMENTAL RESULTS

We conduct comprehensive validation of the proposed method by using five public datasets, i.e., NIREP NA0, LONI LPBA40, IBSR18, CUMC12 and MGH10. All datasets are widely adopted in the literature to evaluate the performance of brain MR image registration algorithms [16], [18]. In pre-processing, skull-stripping has been applied to all images, which are later resampled to the same resolution ($1\times1\times1$ mm$^3$). All images are also processed through bias correction and linearly registered to the MNI152 space by FLIRT in FSL. Note that our pre-processing is consistent with the report in [16] for fair comparison.

Our method is mainly compared to Diffeomorphic Demons [8] and SyN [5], both of which are highly recommended in the large-scale validation in [16]. To quantitatively evaluate the registration performance, we adopt three metrics, i.e., Dice similarity coefficient (DSC), target overlap (TO), and average symmetric surface distance (ASSD) of the annotated tissues or regions-of-interest (ROIs). These three metrics are widely used to quantify registration performance – a higher DSC/TO or lower ASSD usually indicates better registration quality. We note that DSC and ASSD are frequently adopted when evaluated upon large tissues (i.e., GM/WM). For small ROIs, we adopt TO in order to keep consistent with [16].

### A. NIREP Dataset

The NIREP dataset consists of 16 brain MR images, each of which contains GM/WM labeling and 32 small ROIs. In particular, we train all 7 MS-Nets with NIREP only. The trained networks are applied to other datasets in subsequent experiments. To this end, for the evaluation upon the NIREP dataset, we use the leave-two-out strategy. In every test case, there are 14 subjects for training and the rest 2 images act as the pair of the fixed/moving images. Thus, there are 240 test cases for the NIREP dataset in total, from which the evaluation metrics are computed.

TABLE II
REGISTRATION ACCURACY EVALUATED ON THE FOUR DATASETS OF LPBA, CUMC, IBSR, AND MGH.

| Dataset | Method | DSC (%) | | ASSD (mm) | | TO (%) |
|---|---|---|---|---|---|---|
| | | GM | WM | GM | WM | ROIs |
| LPBA | Demons | 76.53±1.88 | 82.55±0.76 | 0.42±0.04 | 0.50±0.09 | 71.12±5.28 (68.93±6.10†) |
| | SyN | 77.74±1.90 | 84.23±0.91 | 0.38±0.04 | 0.45±0.09 | **72.26±5.23** (71.46±5.75†) |
| | Proposed | **82.00±1.50** | **87.86±0.61** | **0.33±0.04** | **0.41±0.09** | 69.48±5.84 |
| CUMC | Demons | 73.03±1.58 | 80.45±0.80 | 0.41±0.05 | 0.50±0.09 | 51.59±15.06 (46.46±15.56†) |
| | SyN | 75.15±1.63 | 82.42±1.11 | 0.37±0.06 | 0.47±0.10 | **52.17±15.11** (51.63±14.88†) |
| | Proposed | **78.62±1.59** | **85.98±0.81** | **0.31±0.04** | **0.42±0.10** | 52.08±14.58 |
| IBSR | Demons | 83.26±2.14 | 78.76±2.61 | 0.47±0.15 | 0.59±0.20 | 51.59±10.51 (46.82±9.89†) |
| | SyN | 84.41±2.40 | 80.28±2.91 | 0.44±0.15 | **0.55±0.19** | **53.96±10.76** (52.81±10.45†) |
| | Proposed | **84.59±3.52** | **81.14±3.88** | **0.35±0.13** | 0.56±0.29 | 49.68±14.58 |
| MGH | Demons | 78.26±1.30 | 81.43±1.29 | 0.40±0.07 | 0.50±0.11 | 56.64±12.45 (52.28±12.97†) |
| | SyN | 80.35±1.39 | 83.77±1.04 | 0.37±0.08 | **0.44±0.08** | **57.04±13.30** (56.83±13.02†) |
| | Proposed | **81.12±2.12** | **86.15±2.25** | **0.29±0.07** | 0.45±0.19 | 53.00±12.05 |

† indicates the results copied from [16]. Note that our reproduced results are clearly better, partially due to the continuous improvement of Demons and SyN in past years.

The evaluation results are summarized in Table I. The DSC scores for our method are 87.06±0.91 (GM) and 89.60±0.72 (WM), both of which are significantly higher than Demons (GM: 81.69±1.02; WM: 84.20±0.80) and SyN (GM: 81.59±2.26; WM: 83.91±2.10). The results of ASSD are similar with DSC, as our method yields significantly superior performances compared to Demons and SyN. Regarding the TO scores of 32 small ROIs, the overall average TO is 70.78±5.00 for our method, compared to 67.39±6.26 for Demons and 66.92±6.96 for SyN. Paired t-tests indicate that, on 27/32 ROIs (except for Left Cingulate Gyrus, Right Cingulate Gyrus, Right Insula Gyrus, Left Temporal Pole and Right Parahippocampal Gyrus), our method performs better than the other two methods (p<0.01). The visualization of the registration results in Fig. 3 shows that our proposed method achieves more accurate surface alignment especially in the regions indicated by red arrows.

The length of the trajectory, or the number of the MS-Nets used, is a critical parameter in our method. We particularly cascade 7 MS-Nets to simplify brain complexity gradually, which is also verified by the experiments on the NIREP dataset. That is, we randomly draw a pair of images from the NIREP dataset for 15 times. Given each drawn image pair, we use the other 14 images in the dataset to train the sequence of MS-Nets. Then, we evaluate the registration quality on the drawn image pair, while different numbers of MS-Nets are used and thus the length of the sequence/trajectory is altered. The results of the DSC scores of GM/WM over the 15 randomly selected pairs for testing are show in Fig. 4. Note that the DSC scores increase rapidly for a short trajectory (i.e., n≤3) and become mostly stable after n=7. To this end, we choose 7 as the optimal number of MS-Nets in all of our experiments.

### B. Other Datasets

With all MS-Nets trained with the NIREP dataset, we apply them directly to the other four datasets. The quantitative results are summarized in Table II.

- The LPBA dataset consists of 40 brain MR images, each of which contains GM/WM labeling and 56 ROIs. In particular, we draw 40×39 pairs of the fixed and moving images from the dataset, which lead to 1560 testing cases in total. The 7 MS-Nets are directly transferred from the NIREP dataset. According to the results in Table II, the DSC scores for our method are 82.00±1.50 (GM) and 87.86±0.61 (WM), both of which are significantly higher than Demons (GM: 76.53±1.88; WM: 82.55±0.76) and SyN (GM: 77.74±1.90; WM: 84.23±0.91). The results of ASSD are similar, as our method performs significantly better than Demons and SyN. The average TO scores of 56 small ROIs, however, is not improved (our method: 69.48±5.84; Demons: 71.12±5.28; SyN: 72.26±5.23). A detailed discussion of TO scores will be provided in the next.

- CUMC consists of 12 brain MR images with 128 ROIs. For evaluation, we conduct 12×11 pairs of registration tasks. While the MS-Nets are directly transferred from the NIREP dataset, the DSC scores for our method are 78.62±1.59 (GM) and 85.98±0.81 (WM), both of which are significantly higher than Demons (GM: 73.03±1.58; WM: 80.45±0.80) and SyN (GM: 75.15±1.63; WM: 82.42±1.11). The results of ASSD are also similar. Regarding the TO scores of the small ROIs, the average TO of our method is mostly comparable (our method: 52.08±14.58; Demons: 51.59±15.06; SyN: 52.17±15.11).

- IBSR consists of 18 brain MR images and 84 ROIs. For the 17×18 pairs of registration tasks, the DSC scores for our method are 84.59±3.52 (GM) and 81.14±3.88 (WM), both of which are significantly higher than Demons (GM: 83.26±2.14; WM: 78.76±2.61) and SyN (GM: 84.41±2.40; WM: 80.28±2.91). The results of ASSD are similar, as our method yields significantly better performance in GM and comparable performance in WM, compared to Demons and SyN. Regarding the TO scores, the average TO of all 84 ROIs is not improved (our method: 49.68±14.58; Demons: 51.59±10.51; SyN: 53.96±10.76).

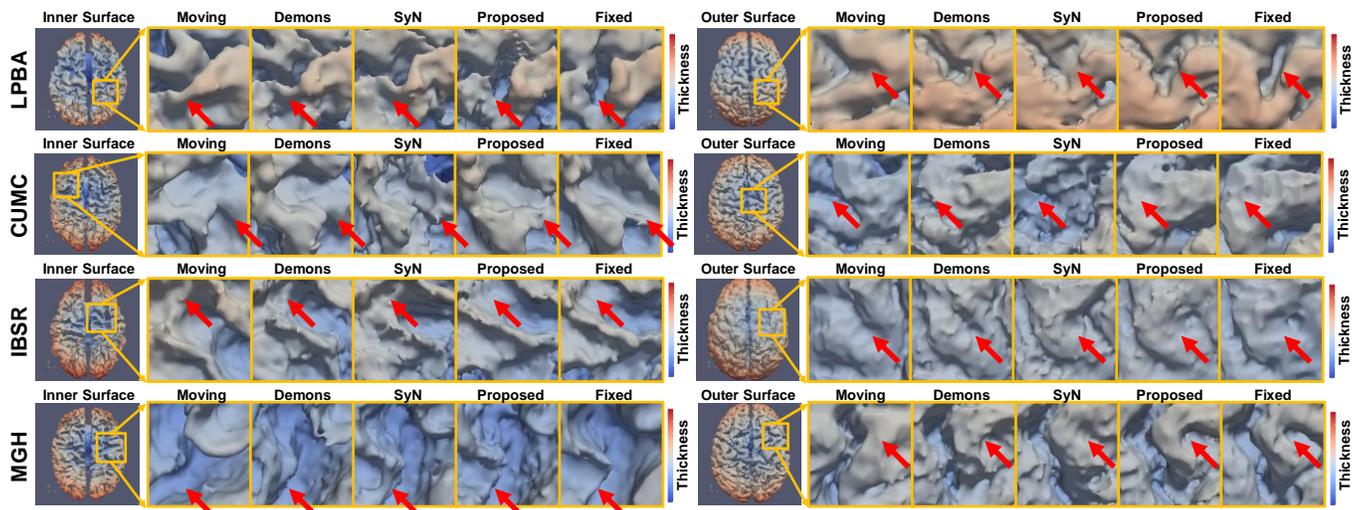

Fig. 5. Visualization of typical registration results by Demons, SyN and our proposed method on the four datasets of LPBA, CUMC, IBSR, and MGH. Our method shows more accurate surface alignment in the regions indicated by the red arrows.

- MGH consists of 10 brain images and 74 ROIs. There are thus 10×9 pairs of registration tasks to evaluate. The DSC scores for our method are 81.12±2.12 (GM) and 86.15±2.25 (WM), both of which are significantly higher than Demons (GM: 78.26±1.30; WM: 81.43±1.29) and SyN (GM: 80.35±1.39; WM: 83.77±1.04). The results of ASSD are similar, as our method yields significantly better performance in GM and comparable performance in WM, compared to Demons and SyN. Regarding the TO scores of 74 ROIs, the average TO of 130 ROIs is not improved (our method: 53.00±12.05; Demons: 56.64±12.45; SyN: 57.04±13.30).

In general, the above quantitative evaluation supports the conclusion that our method can improve the registration accuracy of brain MR images. The disagreement largely comes from the metric of TO scores, while a detailed analysis will be provided later. Moreover, we show the visualization of the typical registration results in Fig. 5. All images in the figure are rendered through their inner and outer cortical surfaces, which are reconstructed from the labeling of GM/WM. We observe that, after being warped through the deformation fields of different methods, the moving image becomes similar with the fixed image. Particularly, the proposed method results in the most accurate alignment of the cortical surfaces, especially in the regions that are highlighted by the red arrows. To this end, we argue that the proposed method can achieve superior registration performance when it is generalized to diverse datasets.

## IV. DISCUSSION

In this paper, we have proposed a novel deep learning based method to guide deformable registration of brain MR images. The MS-Nets simplify morphological complexity of the fixed and the moving images, such that these two images become similar with each other and easy to be registered eventually. Our experiments show superior alignment performance, especially near the cortical surface, attained by our method, compared with the state-of-the-art methods. Moreover, our method has demonstrated its promising generalization capability. While the MS-Nets are trained with a certain dataset (i.e., NIREP), the registration quality on other four datasets are mostly satisfactory.

Note that our method is highly different from deep learning based registration methods, such as VoxelMorph [33]. First, MS-Net is a fully convolution network without pooling, which aims to infer an image with simplified morphology. The output of MS-Net then acts as the target, toward which the input moving image is registered and deformed. Therefore, although the output image of MS-Net reduces appearance complexity with respect to the input image, the actual registration above and the deformed image preserves all information that is inherited from the input image prior to simplification or deformation since the anatomical details are encoded into the estimated deformation field. Second, MS-Net is not designed to estimate the deformation field directly, which is commonly produced through a black-box of deep learning in several recent works. Admittedly, in the future work, we intend to integrate all MS-Nets into a unified network, such that the simplified intermediate images and the deformation pathway between the input images can be generated simultaneously. In this way, we will have a more efficient implementation, while the deformation pathway is clearly tractable.

Although our method has shown its superior performance by visual inspection and by DSC/ASSD scores, one may note that the TO scores of small ROIs of our method are often short of Demons and SyN. A possible reason is because of the inconsistent quality in labeling ROIs, especially in reference to GM/WM boundaries. For example, given an LPBA subject, one may refine the ROIs by intersecting with GM/WM tissue labels [34]. In this way, each ROI can be split into "ROI∩GM" and "ROI ∩ WM". We further investigate the registration quality by computing TO scores in ROI∩GM and ROI∩WM, respectively. The average scores for the four datasets (LBPA, CUMC, IBSR, and MGH) are compared in Fig. 6.

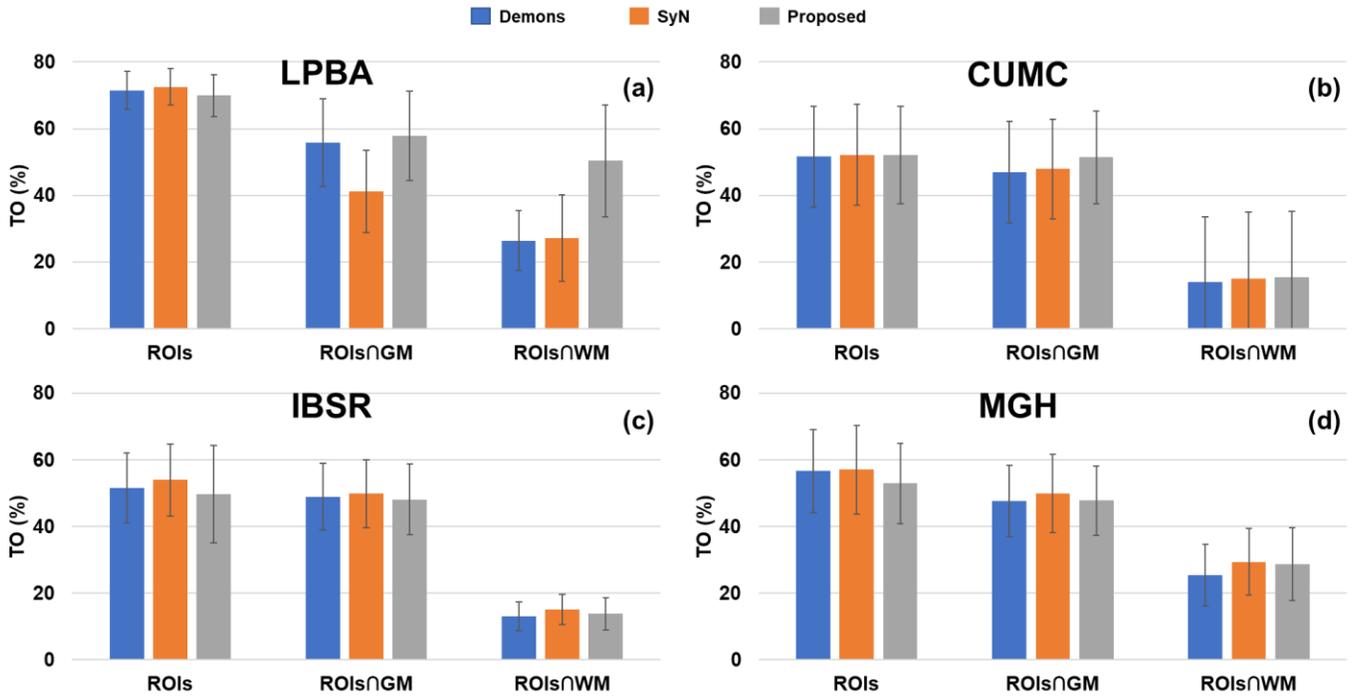

Fig. 6. The TO scores measured from the labeled ROIs and the ROIs intersected by GM/WM. From (a)-(d), the results on four datasets (LPBA, CUMC, IBSR, and MGH) are reported.

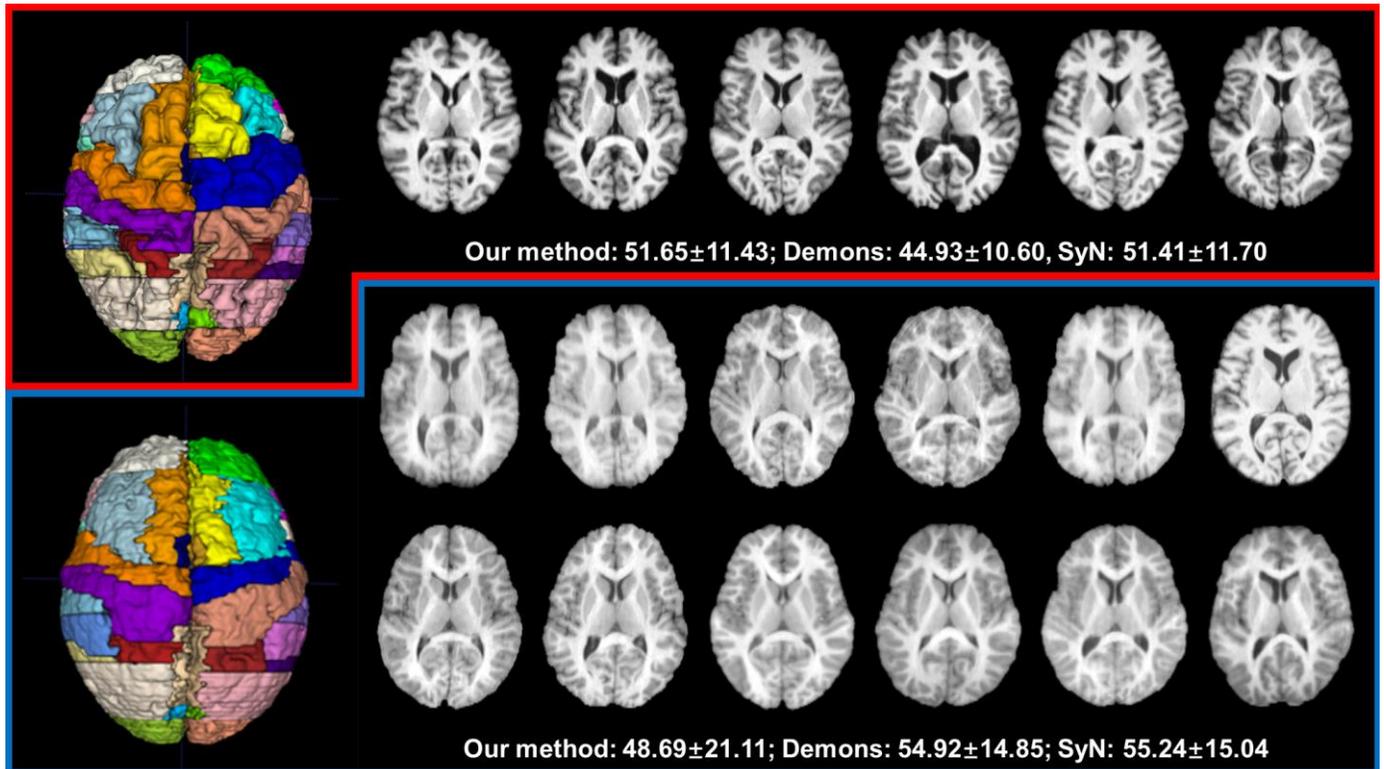

Fig. 7. Visualization of all 18 subjects in the IBSR dataset, as well as the outer cortical surface rendering of the ROIs of two exemplar subjects. The dataset is clearly divided into two groups by appearance. The red group (top) includes 6 subjects with high contrast, while the quality of the 12 subjects in the blue group (bottom) is relatively low. The ROI labeling of the blue group often fails to reveal the subtle gyral and sulcal structures, resulting in an unexpectedly smooth cortical surface labeled by the ROIs.

While the TO scores on the entire ROIs of our method may be slightly lower than the other methods under comparison, it is clear that our method performs better (or in a comparable way) by referring to the splitted ROIs. For example, regarding ROIs ∩ GM, the average TO score for our method on LPBA is 57.91±13.41, compared to 55.83±13.25 (Demons) and 41.15±12.32 (SyN). For CUMC, the scores are 51.44±13.85 (our method), 46.96±15.12 (Demons), and 47.88±14.94 (SyN), respectively. Meanwhile, although our methods are often better in ROIs ∩ WM, we argue that the results of the four datasets might be strongly biased if the ROI labeling within WM is counted in. In particular, the TO scores on ROIs ∩ WM produced by Demons and SyN are very low (usually about 10-30), such that one may challenge whether the boundary of the ROI is determined properly in WM or near the GM/WM interface. In this case, the scores on the small ROIs may not be the proper indicators of registration quality.

In addition to the inner cortical surface between GM and WM, the ambiguity near the outer cortical surface that is partially due to low imaging quality also challenges the reliability of the TO scores. By referring to the visualization of all images in the IBSR dataset in Fig. 7, it is clear to observe that the two groups of images are significantly different in their appearance. The quality of the red group (in the top of the figure) appears better, e.g., with fewer artefacts and clearer details. For image pairs in the red group, the registration performance of our method is much better than other two methods. However, our method fails to compete with other two methods in the blue group by TO scores. We argue that the labeling quality of the ROIs is directly related with the TO scores in this case. Particularly, an example of the blue group shows barely details of the labeled ROIs at the outer cortical surface, which is clearly caused by the low quality of the image itself. A similar observation can also be acquired from the MGH dataset, where the quality of the ROIs labeling might be questionable.

## V. CONCLUSION

In conclusion, this study establishes MS-Net as a powerful and flexible tool to simplify the MR images for deformable registration. The MS-Net provides morphologically simplified images as intermediate guidance, which is also robust to transfer to a new dataset. Our proposed unique method divides the highly complex inter-subject registration task into several easy tasks. Experimental results show superior alignment performance especially near cortical surface compared with state-of-the-art methods on multiple datasets.